\documentclass[journal]{IEEEtran}
\usepackage[utf8]{inputenc}
\usepackage{amsmath,comment}
\usepackage{mathtools}
\usepackage{graphicx} 
\usepackage{amssymb}
\usepackage{dsfont}
\usepackage{float}
\usepackage{soul,color}
\usepackage{hyperref}
\usepackage{todonotes}
\usepackage{physics}
\usepackage{dcolumn}
\usepackage[noadjust]{cite} 
\usepackage{flushend}
\usepackage{subfiles}
\usepackage{multicol}

\DeclareMathOperator*{\argmax}{arg\,max}

\begin{document}

\title{Generalized 
Key-Value 
Memory to Flexibly Adjust Redundancy in Memory-Augmented 
Networks
}



\author{Denis~Kleyko, Geethan~Karunaratne, Jan~M.~Rabaey, Abu~Sebastian, and Abbas~Rahimi
\thanks{
Manuscript received August 6, 2021; revised December 17, 2021; accepted March 10, 2022.
\\
The work of DK was supported by the European Union's Horizon 2020 Programme under the Marie Skłodowska-Curie Individual Fellowship Grant (839179). 
The work was supported in part by the DARPA's AIE HyDDENN Project (DK and JMR) program and by AFOSR FA9550-19-1-0241 (DK). 
GK and AS were partially supported by the European Research Council (ERC) under the European Unions Horizon 2020 research and innovation program (grant agreement
number 682675).
}
\thanks{D. Kleyko is with the Redwood Center for Theoretical Neuroscience at the University of California, Berkeley, CA 94720, USA and also with the Intelligent Systems Lab at Research Institutes of Sweden, 16440 Kista, Sweden. \mbox{E-mail}: \mbox{denkle@berkeley.edu}
}
\thanks{G. Karunaratne, A. Sebastian, and A. Rahimi are with IBM Research - Zurich, 8803 R\"{u}schlikon, Switzerland. \mbox{E-mail}: \mbox{\{kar, ase, abr\}@zurich.ibm.com}
}
\thanks{J. M. Rabaey is with the Department of Electrical Engineering and Computer Sciences at the University of California, Berkeley, CA 94720, USA. \mbox{E-mail}: \mbox{jan\_rabaey@berkeley.edu}
}
}

\markboth{TO APPEAR IN IEEE TRANSACTIONS ON NEURAL NETWORKS AND LEARNING SYSTEMS}%
{Kleyko \MakeLowercase{\textit{et al.}}: 
Generalized Key-Value Memory to Flexibly Adjust Redundancy}

\maketitle

\begin{abstract}
Memory-augmented neural networks enhance a neural network with an external key-value memory whose complexity is typically dominated by the number of support vectors in the key memory.
We propose a generalized key-value memory that decouples its dimension from the number of support vectors by introducing a free parameter that can arbitrarily add or remove redundancy to the key memory representation.
In effect, it provides an additional degree of freedom to flexibly control the trade-off between robustness and the resources required to store and compute the generalized key-value memory.
This is particularly useful for realizing the key memory on in-memory computing hardware where it exploits nonideal, but extremely efficient non-volatile memory devices for dense storage and computation.
Experimental results show that adapting this parameter on demand effectively mitigates up to 44\% nonidealities, at equal accuracy and number of devices, without any need for neural network retraining.
\end{abstract}


\section{Introduction}
The idea of using memory for the neural networks has been widely used since the formulation of long short-term memory~\cite{LSTM}.
Recent approaches to memory-augmented neural networks (MANNs) incorporate an \emph{explicit}  memory into the neural networks as an end-to-end differentiable module~\cite{graves14,graves16,weston15,sukhbaatar15}.
These MANNs are typically applied in knowledge-based reasoning~\cite{graves14,graves16,weston15}, sequential prediction~\cite{sukhbaatar15}, unsupervised learning~\cite{Wu_2018_CVPR,Wu_2018_ECCV}, and few-shot learning tasks~\cite{santoro16,matching_net}. 
All these MANN models commonly expand the explicit memory to be able to handle various tasks and datasets with an increased complexity.
For instance, the size of explicit memory grows linearly with the number of available samples and classes in the few-shot learning tasks~\cite{santoro16,matching_net}, or with the total number of training samples in the unsupervised learning~\cite{Wu_2018_CVPR,Wu_2018_ECCV}.

In the supervised learning tasks, the explicit memory is composed of a key memory for storing and comparing learned patterns, and a value memory for storing labels, that are jointly referred to as a key-value (KV) memory~\cite{weston15}.
The entries in the key memory are not accessed by stating a hard address, but by comparing a query with all the entries, forming soft read and write operations, which involve every individual memory entry.
These extremely memory intensive operations cause a bottleneck when implemented in conventional von Neumann architectures (e.g., CPUs and GPUs), especially for tasks demanding a large number of memory entries.

To address the aforementioned bottleneck, one viable option is to implement the KV memory with emerging non-volatile memory (NVM) devices that offer dense storage as well as in-memory computing capability to efficiently execute the comparison operations at constant time.
For instance, in~\cite{TCAM_NatureElec19} the NVM devices have been arranged as a ternary content addressable memory to perform comparisons inside the key memory.
This structure, however, cannot support widely-used metrics such as cosine similarity. 
Furthermore, practical in-memory computing is challenging due to low computational precision resulting from various sources of 
nonidealities in the NVM devices such as intrinsic randomness, noise, and variability~\cite{Y2020sebastianNatureNano}.
A recent methodology~\cite{KarunaratneHDAugmented2021} addresses these issues by enhancing the key memory representations with robust properties of hyperdimensional computing~\cite{KanervaHyperdimensional2009}, such that the representations can be readily transformed to low-precision (i.e., bipolar or binary) vectors in the key memory while exhibiting robustness against the nonidealities in the NVM-based in-memory computing hardware.

In the few-shot learning tasks, for a given $m$-way $n$-shot problem, the methodology in~\cite{KarunaratneHDAugmented2021} sets the size of the key memory to $[d \times mn]$ where $m$ is the number of classes in the problem, $n$ denotes the number of training samples per class, and $d$ is the dimensionality of support vectors. 
Specifically, the controller neural network in~\cite{KarunaratneHDAugmented2021} assigns a $d$-dimensional support vector to every training sample such that the support vectors for different classes are quasi-orthogonal.
The dimensionality of support vector is typically set to thousands~\cite{KanervaHyperdimensional2009} to generate a  holographic distributed representation that is extremely robust against nonidealities in in-memory computing~\cite{Karunaratne2020}. 
Therefore, the dimensionality of support vectors could be changed to maintain a desired accuracy in the presence of the nonidealities for a fixed $m$-way $n$-shot problem.
However, every time $d$ is changed, the controller neural network should be retrained.
This is impractical as retraining the controller neural network is a costly process. 
Therefore, it is important to consider alternative approaches for regulating the robustness of the KV memory.
%
To address this limitation, we propose a generalization of the KV memory that can be dynamically adapted in the inference phase to deal with any amount of nonidealities.

Thus, the generalization of the KV memory is the main contribution of this brief.
The proposed generalization decouples the dimension of the key memory from $mn$ by introducing a free parameter $r$ for controlling the redundancy such that the dimension becomes $[d \times r]$ instead of $[d \times mn]$. 
During the inference phase, this new parameter $r$ can add or remove redundancy in the representations of the key memory, on demand, without any retraining of the controller neural network.
This results in a fully distributed version of the key memory, which is obtained, by the linear superposition of the outer products between the support vectors and randomized distributed representations of their corresponding class labels.
The empirical investigation of the generalized KV memory demonstrates its flexibility and robustness against noise and NVM nonidealities, e.g., it maintains the noiseless accuracy (obtained in software) of the original KV memory when exposed up to 44\% NVM device nonidealities by adjusting $r$ such that it demands no more NVM devices than the original  KV memory.

The rest of this brief is structured as follows. 
Section~\ref{sec:overview} provides an overview of the original MANN architecture. 
The organization of the proposed generalized KV memory is presented in Section~\ref{sec:generalized}.
Section~\ref{sec:perf} presents performance evaluation of the generalized KV memory.
Section~\ref{sec:conc} concludes the brief.

\section{MANNs overview}
\label{sec:overview}
The MANN architectures combine neural networks with an explicit memory~\cite{graves14,graves16,weston15,sukhbaatar15,Wu_2018_CVPR,Wu_2018_ECCV,santoro16,matching_net,KarunaratneHDAugmented2021}. 
Such an approach exploits meta-learning for performing few-shot learning tasks~\cite{santoro16,matching_net,KarunaratneHDAugmented2021}.
The trained neural network can produce representations of new previously unseen data, which are then written to the explicit memory, so that the memory can be used to, e.g., classify new queries with only a few examples per each class.
The distinctive feature of the MANN architecture in~\cite{KarunaratneHDAugmented2021} is that the neural network is guided to produce support vectors in the form of $d$-dimensional vectors with the properties suitable for hyperdimensional computing~\cite{KanervaHyperdimensional2009, RahimiNanoscalable2017} (also known as vector symbolic architectures~\cite{VSA03, KleykoComputingParadigm2021}).\footnote{
Please consult~\cite{KleykoSurveyVSA2021Part1,KleykoSurveyVSA2021Part2} for a comprehensive survey of hyperdimensional computing/vector symbolic architectures. 
}
The architecture shows that the support vectors produced by the trained neural network can be directed towards robust bipolar or binary representations. 
It was shown to solve the Omniglot~\cite{LakeOmniglot2015} problems, as large as 100-way 5-shot using the in-memory computing hardware.
Specifically, this architecture allows implementation of the binary key memory on 256,000 ($5 \times 100 \times 512$) noisy phase-change memory (PCM) devices, performing highly efficient analog in-memory computation, with less than 2.7\% accuracy drop compared to the 32-bit real-valued memory in software for the largest problem ever-tried on Omniglot~\cite{KarunaratneHDAugmented2021}.

As follows from the above, conceptually the architecture can be divided into two parts: the controller (i.e., the neural network) and the explicit memory.
This brief is devoted to the organization of the explicit memory during the inference phase.\footnote{
It is important to note that the proposed approach does not require any modification of the training of the controller described in~\cite{KarunaratneHDAugmented2021}.
}
The explicit memory is split into two parts: the key memory and the value memory (hence, the KV memory).
In the original formulation, the key memory (denoted as $\mathbf{K}$) stores $mn$  $d$-dimensional\footnote{Following~\cite{KarunaratneHDAugmented2021}, $d$ is set to $512$ for the experiments in this brief.} support vectors produced by the controller for a given $m$-way $n$-shot problem so $\mathbf{K} \in [d \times mn]$.
For the inference phase with PCM devices, every $d$-dimensional support vector is quantized to a binary or bipolar vector.
The value memory (denoted as $\mathbf{V}$) stores one-hot encodings of the class labels corresponding to the support vectors in $\mathbf{K}$, so $\mathbf{V} \in [m \times mn]$.
Here, it is important to emphasize that despite the fact that the KV memory contains the  holographic distributed representations (i.e., the support vectors), the organization of the memory is local because the vectors $\mathbf{K}_i$ and $\mathbf{V}_i$ in the corresponding parts of the KV memory can be identified with a particular sample of the training data. 

During the inference phase, the query vector $\mathbf{q} \in [d \times 1]$ is used as an input to the key memory where the main step is to compute the similarity between $\mathbf{q}$ and the support vectors $\mathbf{K}_i$ using the dot product (denoted as $\boldsymbol\alpha$) as the similarity measure:
\begin{equation}
\boldsymbol\alpha = \mathbf{K}^{\top} \mathbf{q},
\end{equation}
where $\alpha_i$ contains the similarity between the query and $i$th support vector.
Note that when the support vectors in the key memory are normalized to the same norm, the dot products in $\boldsymbol\alpha$ are proportional to the corresponding cosine similarities. 

Next, the dot products can be modified with some sharpening function (denoted as $\sigma(\cdot)$):
\begin{equation}
\boldsymbol\gamma = \sigma(\boldsymbol\alpha).
\end{equation}
The sharpened similarity scores are used to compute the accumulated scores for each class as:
\begin{equation}
\label{eq:sharp}
\mathbf{s} = \mathbf{V} \boldsymbol\gamma,
\end{equation}
where $s_j$ contains the score for $j$th class ($1 \leq j \leq m $).
The prediction is chosen to be the class with the highest accumulated score: $\underset{j}{\argmax} \: s_j$.

It is worth noting that the above inference procedure is a special case of the $k$-nearest neighbor classifier with distance-weighted voting where $k=mn$ and the weight for $i$th training sample (i.e., neighbor) corresponds to $\gamma_i$.
This observation suggests that it is worth exploring a \emph{fully distributed} organization of the KV memory where there is no \emph{local} correspondence between the entries of the key memory and the support vectors.
Such a fully distributed organization can be achieved using, e.g., a linear distributed memory with the outer product learning rule~\cite{MizrajiContext1989,Frady2019robust}.
The distributed organization of the KV memory allows achieving similar functionality for the inference phase while providing an additional degree of freedom, which plays an important role in controlling the trade-off between the robustness of the key memory and the resources required to store it. 

\section{Generalized key-value memory}
\label{sec:generalized}

As highlighted in~\cite{KarunaratneHDAugmented2021}, an important advantage of the architecture is that the key memory can be mapped to PCM devices for analog in-memory computing, which has shown to significantly improve the energy efficiency of the inference procedure compared to a digital design.
Note, however, that in the original formulation, the dimension of the key memory is $\mathbf{K} \in [d \times mn]$, i.e., assuming that $d$ is fixed, the dimension of the key memory is determined by the number of support vectors in a given $m$-way $n$-shot problem.
This dependency makes the key memory rigid in the sense that there is no possibility to control the dimension of the key memory other than changing $d$, which demands retraining the controller neural network, once $m$ and $n$ are fixed.
Therefore, it is important to consider a generalization of the KV memory allowing to decouple the dimension of the key memory from $m$ and $n$ by using a free parameter $r$ so the dimension becomes $[d \times r]$ instead of $[d \times mn]$. 

The decoupling is achieved by changing the organization of the KV memory from the local one to a fully distributed one using the principles of forming context-dependent associations in linear distributed associative memories~\cite{MizrajiContext1989,FrolovTime2006,GritsenkoAMSurvey2017}.
To reformulate the KV memory in terms of the context-dependent associations, we need to first define a new value memory 
$\prescript{\text{(d)}}{}{\mathbf{L}} \in [r \times m]$, where $\prescript{\text{(d)}}{}{\mathbf{L}}_j$ is an $r$-dimensional vector representing the label of $j$th class ($1 \leq j \leq m $).
We will discuss the options for choosing $\prescript{\text{(d)}}{}{\mathbf{L}}_j$ in the next section.
Once the class labels' representations are defined, the real-valued support vectors $\mathbf{K}_i$ and their corresponding $\prescript{\text{(d)}}{}{\mathbf{L}}_j$ are used to create the distributed version of the key memory (denoted as $\prescript{\text{(d)}}{}{\mathbf{K}}$) using the outer product learning rule: 
\begin{equation}
\prescript{\text{(d)}}{}{\mathbf{K}} = \sum_{i=1}^{mn} \prescript{\text{(d)}}{}{\mathbf{L}}_{c(i)} \mathbf{K}_i^{\top},
\label{eq:mem:ass}
\end{equation}
where $c(i)$ denotes the class index of $i$th support vector.
Thus, the distributed version of the key memory is the linear superposition of the outer products of the support vectors and the representations of their class labels,\footnote{
While here we do not go into the details of comparing the computational costs of the two considered approaches to the organization of the KV memory, it is worth noting that both of them incur certain computation costs when it comes to adding new support vectors to the key memory.  
In the case of the original KV memory, the cost is in the allocation of space for a new support vector and writing the new vector to the key memory; 
while in the case of the generalized KV memory, the cost is in computing the outer product between the support vector and its label vector as well as in incrementing the key memory with the outer product result.
}
therefore, $\prescript{\text{(d)}}{}{\mathbf{K}} \in [r \times d]$.
Since the distributed version of the key memory in (\ref{eq:mem:ass}) does not strictly depend on $mn$, we refer to $\prescript{\text{(d)}}{}{\mathbf{K}}$ and $\prescript{\text{(d)}}{}{\mathbf{L}}$ as the generalized KV memory.

The inference procedure with the generalized KV memory is very similar to the original one. 
For a given query, $\mathbf{q}$, the dot product between $\prescript{\text{(d)}}{}{\mathbf{K}}$ and $\mathbf{q}$, which is computed as:
\begin{equation}
\boldsymbol\gamma  = \boldsymbol\alpha = \prescript{\text{(d)}}{}{\mathbf{K}} \mathbf{q},
\end{equation}
produces an $r$-dimensional vector, which is the weighted superposition of class labels' representations.
This vector $\boldsymbol\alpha$  can be seen as a sharpened vector $\boldsymbol\gamma$ assuming that an identity function is used as the sharpening function.
Finally, the value memory is used to measure the scores of each class as:
\begin{equation}
\mathbf{s}  = \prescript{\text{(d)}}{}{\mathbf{L}}^{\top} \boldsymbol\gamma.
\end{equation}
\noindent
As in the original KV memory, the result is an $m$-dimensional vector where $j$th component contains the accumulated score for the corresponding class so the prediction is chosen as before: $\underset{j}{\argmax} \: s_j$.

Similar to the original key memory $\mathbf{K}$, $\prescript{\text{(d)}}{}{\mathbf{K}}$ can be bipolarized using the component-wise $\text{sign}(\cdot)$ function:
\begin{equation}
    \prescript{\text{(d)}}{}{\hat{\mathbf{K}}}=\text{sign}(\prescript{\text{(d)}}{}{\mathbf{K}}).
\end{equation}
Obviously, the bipolar version $\prescript{\text{(d)}}{}{\hat{\mathbf{K}}}$ can be transformed to the binary version.

\begin{figure*}[tb]
\centering
\includegraphics[width=2.00\columnwidth]{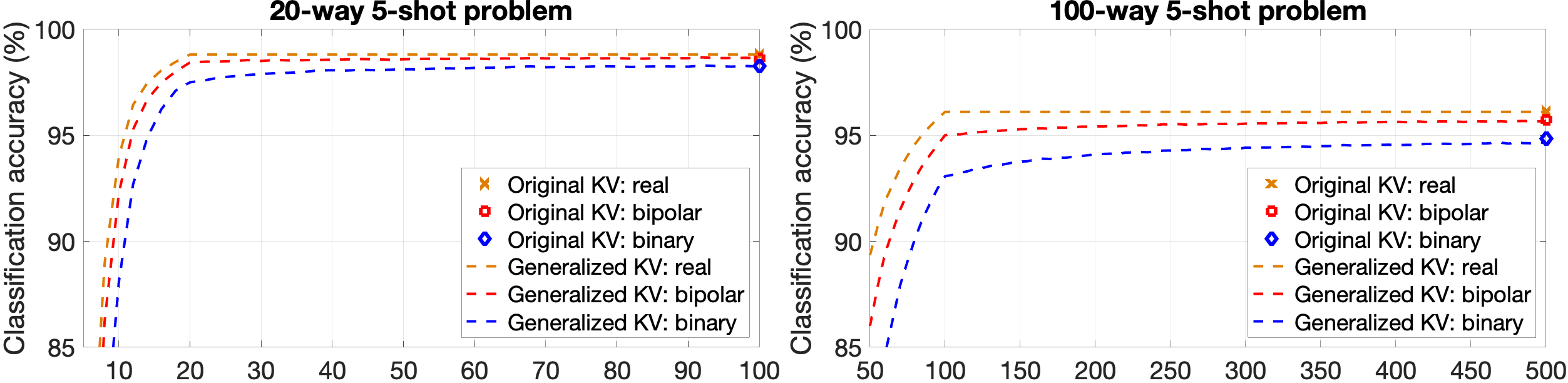}
\caption{
Average classification accuracy against the dimensionality of representations of the class labels ($r$).
The results are shown for the real, bipolar, and binary variants of the key memory and query vectors, in software without any noise.
The results are averaged over $1,000$ problems randomly chosen from the test data.
}
\label{memory:noiseless}
\end{figure*}

\section{Evaluation of the generalized key-value memory}
\label{sec:perf}
Here, we discuss when would it be beneficial to use the generalized key-memory instead of the original KV memory.
We suggest the following two modes: 
\begin{enumerate}
    \item For compression when there is a little or no noise. For the problems where $n>1$, the generalized KV memory effectively removes the redundancy, and can achieve the classification accuracy on a par to the original KV memory for $r<<mn$. 

    \item For increasing robustness at very low signal-to-noise ratio (SNR) conditions. The generalized KV memory flexibly increases the redundancy by allocating more resources to the key memory so that $r>mn$ that results in improved robustness to noise and nonidealities compared to the original KV memory, which does not have a mechanism to regulate the redundancy of the key memory.
    
\end{enumerate}

Both of the aforementioned modes are related to each other since usually it is necessary to tradeoff between the achieved accuracy and required resources in the presence of noise and nonidealities but to make the points clear, we isolate the two. 
In the following, we present the results of three sets of experiments to illustrate these modes.
The first mode is investigated in Section~\ref{sec:perf:compr} while the second one is studied in Sections~\ref{sec:perf:white} and~\ref{sec:perf:PCM}.

We use the data from~\cite{KarunaratneHDAugmented2021} in the form of $512$-dimensional real-valued vectors obtained from the trained controller using the images in the test set of the Omniglot dataset.
It includes $659$ classes with $20$ samples per each class.
These data were used to perform the experiments below. 

\begin{figure*}[tb]
\vspace{19px}
\centering
\includegraphics[width=2.00\columnwidth]{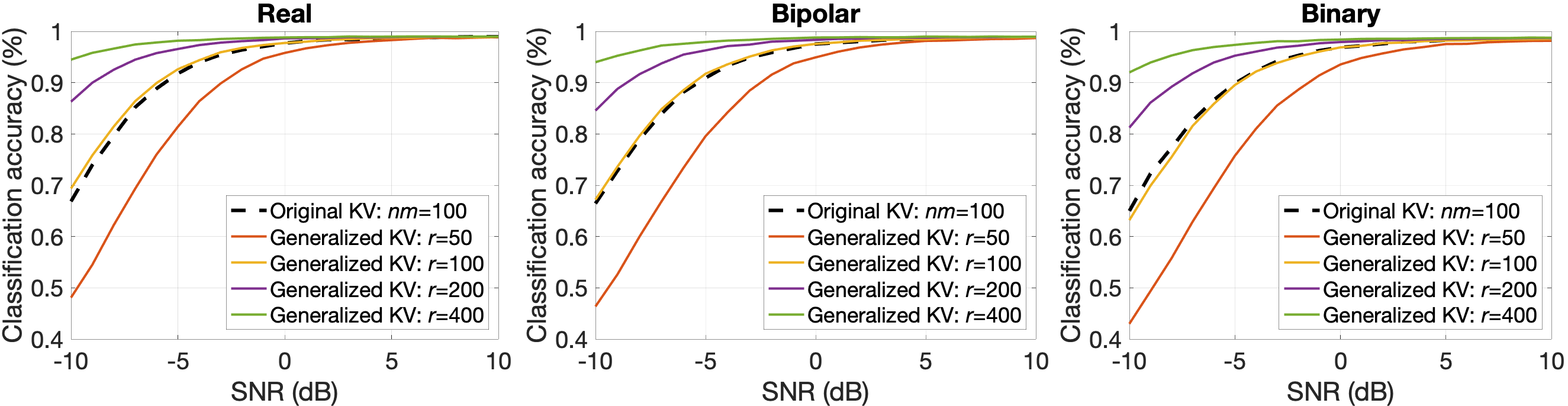}
\caption{
Average classification accuracy against SNR of $\boldsymbol\alpha$.
The $20$-way $5$-shot problem is used with fixed $nm=100$ while $r<nm$ or $r\geq nm$.
Panels correspond to real, bipolar, and binary variants of the key memory and query vectors in software.
The results are averaged over $100$ problems randomly chosen from the test data.
}
\vspace{19px}
\label{memory:noisy}
\end{figure*}

Recall, that the first step in the generalized KV memory is to populate the value memory $\prescript{\text{(d)}}{}{\mathbf{L}} \in [r \times m]$.
There are several options to do so. 
The most obvious choice is to generate $\prescript{\text{(d)}}{}{\mathbf{L}}$ randomly so that each class label is represented by a random vector $\prescript{\text{(d)}}{}{\mathbf{L}}_j$.
This, however, is not the best choice since the random vectors are only approximately orthogonal, hence, there will be some cross-talk noise between different $\prescript{\text{(d)}}{}{\mathbf{L}}_j$ in $\prescript{\text{(d)}}{}{\mathbf{K}}$, which would negatively affect the classification accuracy and robustness to noise and nonidealities. 
Therefore, in the experiments below we use random orthogonal matrices to form the value memory $\prescript{\text{(d)}}{}{\mathbf{L}}$ when $r\geq m$, or whitened random matrices when $r < m$.
In the former case, an orthogonal matrix can be formed by applying the QR decomposition to a random matrix generated from the standard normal distribution.
The fact that values of $\prescript{\text{(d)}}{}{\mathbf{L}}$ are real-valued should not be discouraging as it is meant to be implemented in software.
In case if $\prescript{\text{(d)}}{}{\mathbf{L}}$ is also desired to be binary/bipolar, one could use, e.g., Walsh codes to form $\prescript{\text{(d)}}{}{\mathbf{L}}$.

\subsection{Compression in Noiseless Condition}
\label{sec:perf:compr}
In the first experiment, we evaluate the trade-off between the classification accuracy of the generalized KV memory and its dimension, which is controlled by $r$. 
Fig.~\ref{memory:noiseless} presents the average classification accuracy against $r$ for two problems: $20$-way $5$-shot (left panel) and $100$-way $5$-shot (right panel).
The markers depict the baselines obtained with the original KV memories\footnote{
All the experimental results for the original KV memory were obtained using the identity function as the sharpening function in (\ref{eq:sharp}).
}
Recall, that the dimension of the original key memory is $mn$, i.e., $100$ and $500$ for the considered problems, respectively.
The dashed lines correspond to the results of the generalized KV memories.
For both problems, to provide the intuition for the dimensions of the original key memories, $r$ was limited to $mn$.
It is clear that for both problems, the accuracy of the generalized KV memories approached the accuracy of the original KV memories with values of $r$ being smaller than $mn$.
For example, for the $20$-way $5$-shot problem, the generalized KV memory reached 95\% of the accuracy of the original KV memory by using $r$ of $10$, $12$, and $14$ for real, bipolar, and binary key memory, respectively; these correspond to $mn/r$ ratio, hence, memory saving of $10.0\times$, $8.3\times$, and $7.1\times$.
For the $100$-way $5$-shot problem, the corresponding values of $r$ were $60$, $70$, and $80$, with memory savings of $8.3\times$, $7.1\times$, and $6.3\times$, respectively.
These results demonstrate the possibility of compressing the KV memory without sacrificing most of the classification accuracy. 

\begin{figure*}[tb]
\centering
\includegraphics[width=2.00\columnwidth]{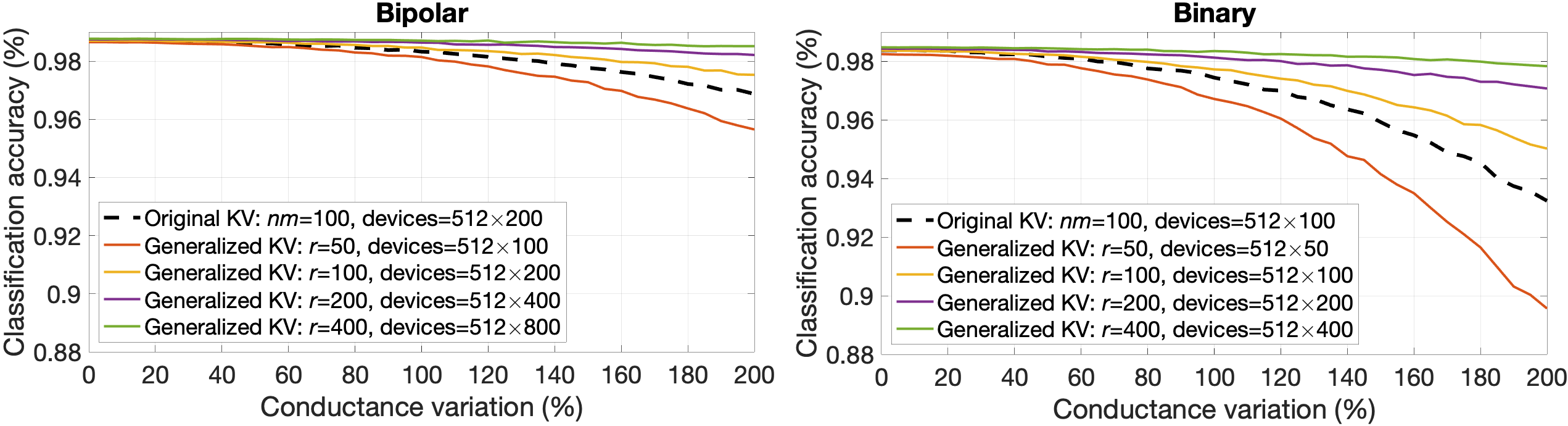}
\caption{
Average classification accuracy against the conductance variations in the PCM model for $20$-way $5$-shot problem.
Panels correspond to the bipolar and the binary variants of the key memory mapped to PCM devices.
The results were averaged over $1,000$ problems randomly chosen from the test data.
}
\label{memory:noisy:pcm:20}
\end{figure*}

\begin{figure*}[tb]
\vspace{20px}
\centering
\includegraphics[width=2.00\columnwidth]{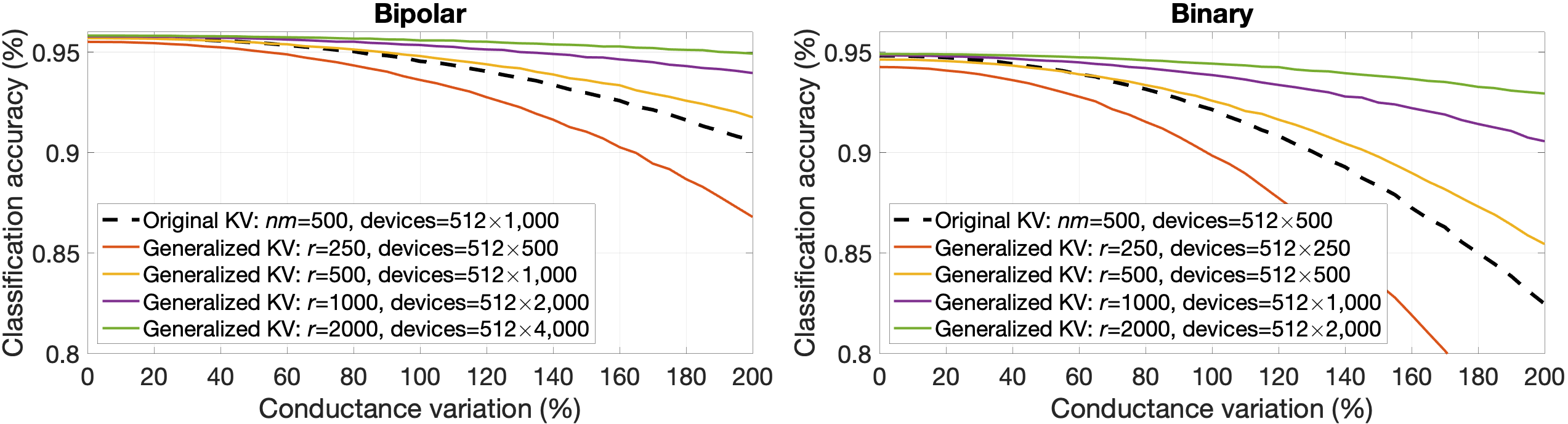}
\caption{
Average classification accuracy against the conductance variations in the PCM model for $100$-way $5$-shot problems.
Panels correspond to the bipolar and the binary variants of the key memory mapped to PCM devices.
The results were averaged over $1,000$ problems randomly chosen from the test data.
}
\vspace{22px}
\label{memory:noisy:pcm:100}
\end{figure*}

\subsection{Robustness in the Presence of White Noise}
\label{sec:perf:white}

The previous experiment did not take into account the fact that a hardware implementation of the key memory might return a noisy version of $\boldsymbol\gamma$ as the result of computing the dot products.
Therefore, in the second experiment, we measure the classification accuracy against the white noise added to $\boldsymbol\gamma$.
The experiment is conducted with $20$-way $5$-shot problem using all three variants of the key memory: real, bipolar, and binary. 
The dimension of the original key memory was fixed to $nm=100$, while for the generalized KV memory flexibly sets four different values with $r \in \{50, 100, 150, 200 \}$ without any need to retrain the controller.

Fig.~\ref{memory:noisy} presents the results. 
As expected, the low SNR values reduced the classification accuracy.
At the same time, it is clear that the dimension of the key memory has an effect on the robustness to noise. 
The generalized KV memory with the lowest $r$ ($r=50$) demonstrated the worst performance for the low SNR values.
When $r=mn$, both memory types performed very close to each other independent of the SNR values.
However, since the generalized KV memory can control $r$, it can be set to $r>mn$. 
Thus, the generalized KV memory can be flexibly switched to the second mode with a larger amount of redundancy to provide robustness to extremely low SNR values.
For example, $r=400$ exhibits a very graceful accuracy degradation (e.g., an accuracy of $>90\%$ at SNR=-10\,dB). 
These results provide the evidence that subject to sufficient dimensionality of $r$, the generalized KV memory can be designed to operate at very low SNR.

\begin{figure*}[tb]
\centering
\includegraphics[width=2.00\columnwidth]{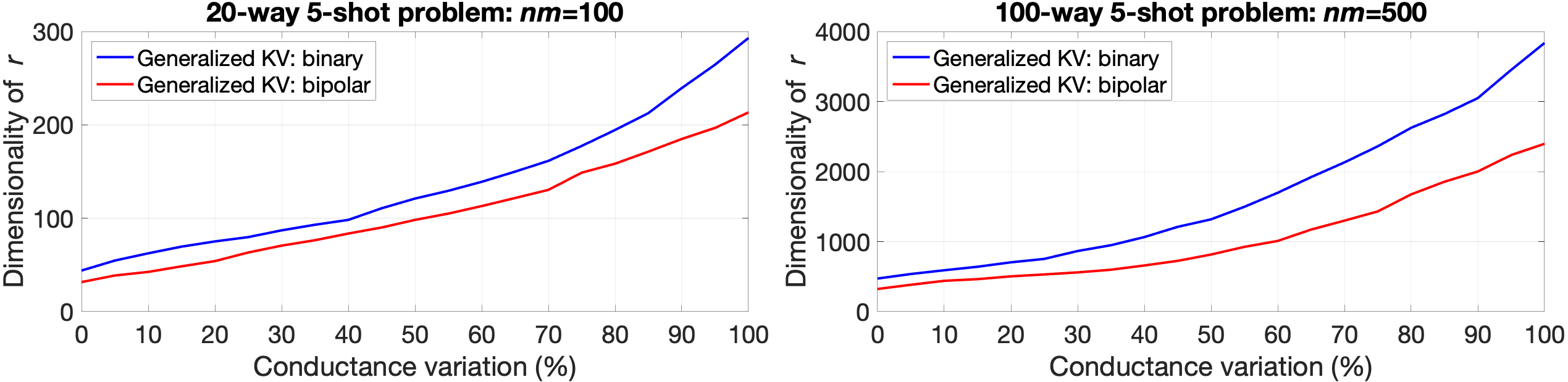}
\caption{
Average dimensionality of $r$ against the conductance variation in the PCM model required to maintain iso-accuracies of the corresponding variants of the original noiseless KV memory. 
Panels correspond to $20$-way $5$-shot and $100$-way $5$-shot problems, respectively. 
The results are averaged over $1,000$ problems randomly chosen from the test data.
}
\vspace{23px}
\label{memory:noisy:pcm:dimen}
\end{figure*}

\subsection{Robustness in the Presence of PCM Nonidealities}
\label{sec:perf:PCM}

It is important to note that the noise and nonidealities present in the PCM devices are not well-described by white noise. It has been shown that there are three major components to PCM noise. These include a) a programming noise component which is modeled as multiplicative Gaussian noise, b) drift noise component which models $1/f$ noise as a Gaussian random exponent with respect to time c) read noise component which is modeled as additive Gaussian noise (see Supplementary notes in \cite{KarunaratneHDAugmented2021}). Of these, the programming noise variability can be directly controlled by employing an iterative programming scheme whereas drift and noise components are controlled by external conditions such as temperature or the internal conductance state, over which there is less controllability. 
Therefore, we perform the last set of experiments using the model of a temporal evolution of conductance $G(t)$ of a single PCM device~\cite{KarunaratneHDAugmented2021}:
\begin{equation}
G(t) = \mathcal{N}(0, \tilde{G}_r^2) + (G_0 \cdot  \mathcal{N}(1, \tilde{G}_p^2)) \cdot  t^{- \nu  \cdot \mathcal{N}(1, \tilde{\nu}^2))},
\label{eq:PCM:model}
\end{equation}
\noindent
where $\mathcal{N}(\mu, \sigma^2)$ denotes a normal distribution; $t$ is the time since programming (assumed to be $20$s), 
$G_0$ is the mean conductance at $t = 1$s (measured: $G_0=22.8 \times 10^{-6}$S),
$\nu$ is the mean drift exponent (measured: $\nu=0.0598$);
$\tilde{G}_r^2$ , $\tilde{G}_p^2$, and $\tilde{\nu}^2$ represent the variation in additive read noise (measured: $\tilde{G}_r=0.496 \times 10^{-6}$S), conductance variation (programming noise; measured: $\tilde{G}_p=31.7$\%), and drift variation (measured: $\tilde{\nu}=9.07$\%), respectively.
Please refer to ``PCM model and simulations'' subsection in ``Method'' section in~\cite{KarunaratneHDAugmented2021} for the additional details of the model.
We perform the experiments by varying the relative conductance variation (i.e., $\tilde{G}_p$) while keeping all other parameters fixed.

Fig.~\ref{memory:noisy:pcm:20} and~\ref{memory:noisy:pcm:100} present the average classification accuracy against the conductance variation in the PCM model for $20$-way $5$-shot and $100$-way $5$-shot problems, respectively.\footnote{
The diligent readers are kindly referred to the Supplementary Material that provides an additional experimental evaluation to further justify the proposed approach.
}
The dimension of the original key memories are fixed to $nm$, while for the generalized KV memories different values of $r$ are used: $\{50, 100, 150, 200 \}$ and $\{250, 500, 1000, 2000 \}$, respectively.
Note that the PCM implementation of the bipolar variants requires two devices per dimension. 

The results for the original KV memory are consistent with the ones reported in~\cite{KarunaratneHDAugmented2021} in the sense that the bipolar variant is more robust against the conductance variations.
Also, similar to the results in Fig.~\ref{memory:noisy}, for both problems the generalized KV memory with the lowest $r$ ($r=50$ and $r=250$, respectively) consistently demonstrates the lowest accuracy amongst all depicted.
Importantly, the generalized KV memory performs better than the original one for high conductance variations ($>80$\%) when $r=nm$ (i.e., having the same number of devices).
This can be attributed to the fact that the local organization is more brittle to errors than the distributed one, which is a well-known advantage of distributed representations~\cite{FradyCapacity2018}. 
%
%
Finally, the robustness to the conductance variation can be increased further by increasing $r$, and naturally, the largest values of $r$ ($r=400$ and $r=2,000$, respectively) demonstrate the least accuracy degradation even at the very high conductance variation.
In particular, for $\tilde{G}_p=200$\% compared to $\tilde{G}_p=0$\%  the accuracy decreases by only $0.26$\% ($r=400$) \& $0.90$\% ($r=2,000$) and $0.64$\% ($r=400$) \& $1.97$\% ($r=2,000$) for the bipolar and binary variants, respectively.

\subsubsection*{Iso-accuracy Generalized KV Memory at Conductance Variations}
In the previous experiment, the generalized KV memory improves the robustness to conductance variations by increasing $r$.
Hence, the next step is to investigate whether the generalized KV memory can achieve the iso-accuracy of the original KV memory without any noise and nonidealities as in Fig.~\ref{memory:noiseless}.
%
%
The average values of $r$ providing the iso-accuracy for a range of conductance variations are depicted in Fig.~\ref{memory:noisy:pcm:dimen}.

Considering the 20-way 5-shot problem with the binary vectors, the generalized KV memory can maintain the original accuracy of its noiseless software variant when experiencing up to 44\% variations in the PCM hardware, yet using the same number of devices ($r=mn$).   
Since in the hardware implementation, the ``set'' conductance was fixed at 38~{\textmu}S, 44\% conductance variation translates into a standard deviation of the ``set'' conductance distribution of 16~{\textmu}S. 
%
For the larger amount of variations, increased $r$ can guarantee the iso-accuracy; as expected, for the bipolar KV memories $r$ grows slower than that of the binary ones.
Similar trend is observed for the large 100-way 5-shot problem. 
As shown, it is possible to achieve the iso-accuracy even for extremely high conductance variations, which confirms the general nature of the proposed KV memory, which allows trading-off additional hardware resources for the robustness of the classification accuracy.

\section{Conclusion}
\label{sec:conc}

This brief presented the generalized KV memory for MANNs. 
The proposed approach is based on the two key ideas. 
First, the fact that it is not necessary to use one-hot encodings for the value memory; randomized distributed representations can be used instead.
Second, the organization of the key memory can be changed from the local one (storing individual support vectors) to the distributed one with the outer product learning rule.

The empirical evaluation demonstrated the flexibility of the
generalized KV memory.
In the noiseless conditions, it achieves the classification accuracy on a par with the original KV memory using a fraction of dimensions required by the original key memory.
Alternatively, for very harsh conditions the generalized KV memory can easily adjust the memory redundancy to tolerate extreme amounts of noise and nonidealities, which is an attractive feature for implementing the KV memory on emerging computational NVM devices.

\newpage

\onecolumn

\begin{center}
{\Large
\textbf{
Supplementary Materials: Additional Evaluation
}
}
\end{center}




\renewcommand\thefigure{S.\arabic{figure}}    
\setcounter{figure}{0}

\begin{figure}[ht]
\begin{center}
\includegraphics[width=1.0\columnwidth]{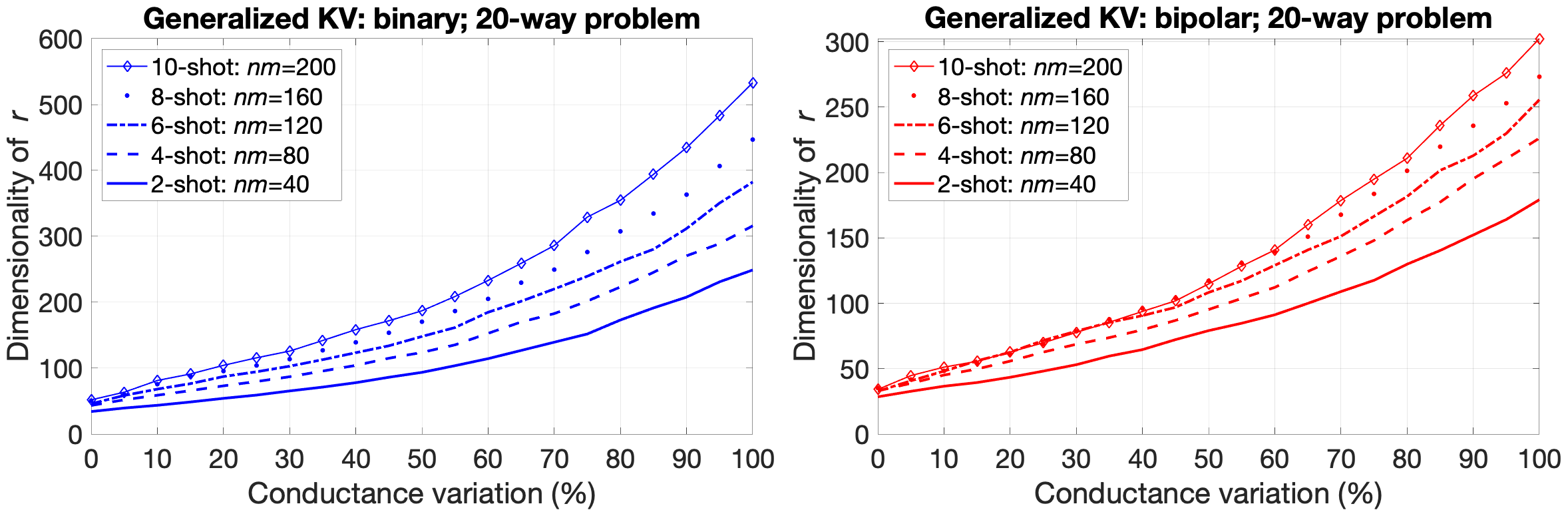}
\caption{
Average dimensionality of $r$ against the conductance variation in the PCM model required to maintain iso-accuracies of the corresponding variants of the original noiseless KV memory. 
Panels correspond to binary and bipolar variants of the generalized KV memory, respectively. 
The number of ways was fixed to $m=20$ while the number of shots was $n \in \{2,4,6,8,10\}$.
The results are averaged over $3,000$ problems randomly chosen from the test data.
} 
\label{fig:scale:n} 
\end{center}
\end{figure}

\begin{figure}[ht]
\begin{center}
\includegraphics[width=1.0\columnwidth]{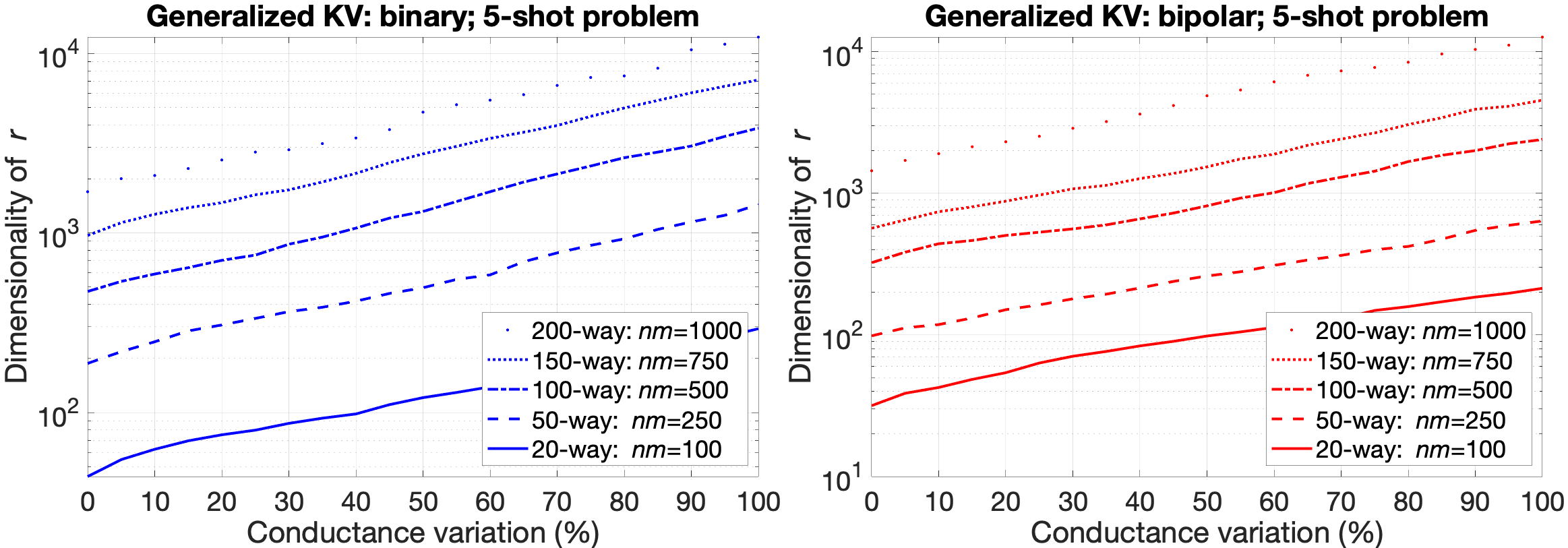}
\caption{
Average dimensionality of $r$ against the conductance variation in the PCM model required to maintain iso-accuracies of the corresponding variants of the original noiseless KV memory. 
Panels correspond to binary and bipolar variants of the generalized KV memory, respectively. 
The number of shots was fixed to $n=5$ while the number of ways was $m \in \{20, 50, 100, 150, 200\}$.
The results are averaged over $1,000$ problems randomly chosen from the test data except for the $200$-way $5$-shot problems that are much more computationally demanding so only $200$ and $700$ problems were simulated for binary and bipolar variants, respectively. 
} 
\label{fig:scale:m} 
\end{center}
\end{figure}

\noindent
The goal of this Supplementary materials is to provide the reader with the additional experimental evaluation to further justify the goodness of the proposed generalized key-value (KV) memory.  
Due to the space limitation, in the main text we have not provided the experiments to demonstrate how the proposed approach scales with respect to the size of the problems -- determined by $mn$. 
In order to make such a demonstration we have run the experiments for fixed $m$ and varying $n$ (see Fig.~\ref{fig:scale:n}) and then for fixed $n$ and varying $m$ (see Fig.~\ref{fig:scale:m}).
In both experiments, we can see that the proposed approach is able to achieve the iso-accuracies at the expense of the increased $r$, which was the main feature of interest for these experiments. 
We also see that for a fixed conductance variation, larger problems (for either increased $n$ or $m$) require larger $r$. 
This is expected, in both cases since in both cases $mn$ is increasing and it is natural to expect that the required $r$ would have increased as well.


\newpage
\bibliographystyle{IEEEtran} 

\bibliography{references}

\end{document}